\pdfoutput=1

\documentclass[11pt]{article}

\usepackage{acl}

\usepackage{times}
\usepackage{latexsym}

\usepackage[T1]{fontenc}

\usepackage[utf8]{inputenc}

\usepackage{microtype}
\usepackage{booktabs}

\usepackage{csquotes}
\usepackage{todonotes}
\usepackage{xspace}
\usepackage{amsmath}
\usepackage{amssymb}
\usepackage{float}
\usepackage{wrapfig}
\usepackage{lipsum}
\usepackage{xcolor,colortbl}
\usepackage{fdsymbol}
\usepackage{multirow}
\usepackage{rotating}
\usepackage{algorithm,algpseudocode}
\usepackage{hyperref}

\usepackage[linewidth=1pt]{mdframed}

\definecolor{green1}{HTML}{d1f3d3}
\definecolor{red1}{HTML}{c20a0a}

\newcommand\rev[1]{\textcolor{black}{#1}}

\newcommand{\squad}{SQuAD\xspace}
\newcommand{\ftqa}{Fairytale QA\xspace}

\newcommand{\bleu}{BLEU-4\xspace}
\newcommand{\rougel}{ROUGE-L\xspace}
\newcommand{\ngram}{$n$-gram\xspace}
\newcommand{\gpt}{GPT-3\xspace}

\newcommand{\Msmax}{$M_{\overline{s}}$\xspace}
\newcommand{\Msmean}{$M_s$\xspace}
\newcommand{\Msmin}{$M_{\underline{s}}$\xspace}
\newcommand{\Mgrdy}{$M_g$\xspace}

\newcommand{\code}[1]{\texttt{#1}}

\newcommand{\token}[1]{\textcolor{darkblue}{\textbf{\small{\code{#1}}}}}

\DeclareMathOperator*{\argmax}{arg\,max}

\title{Selecting Better Samples from Pre-trained LLMs: \\A Case Study on Question Generation}

\author{Xingdi Yuan$^\clubsuit$\thanks{\:\:\:\:Equal contribution.} \:\:\:\: Tong Wang$^{\clubsuit}$\footnotemark[1] \:\:\:\: Yen-Hsiang Wang$^{\diamondsuit}$\footnotemark[1] \\  \textbf{Emery Fine$^{\clubsuit}$ \:\:\:\: Rania Abdelghani$^\spadesuit$ \:\:\:\: Pauline Lucas$^{\spadesuit}$}\\
\textbf{H\'{e}l\`{e}ne Sauz\'{e}on$^{\spadesuit}$ \:\:\:\: Pierre-Yves Oudeyer$^\spadesuit$}\\
$^\clubsuit$Microsoft Research, Montr\'{e}al 
$^\diamondsuit$ National Chung Hsing University 
$^\spadesuit$INRIA \\
{ \tt \{eric.yuan, tong.wang\}@microsoft.com}\\
{ \tt heliart@smail.nchu.edu.tw}\\
}

\begin{document}
\maketitle
\begin{abstract}

Large Language Models (LLMs) have in recent years demonstrated impressive prowess in natural language generation.
A common practice to improve generation diversity is to sample multiple outputs from the model.
However, there lacks a simple and robust way of selecting the best output from these stochastic samples.
As a case study framed in the context of question generation, we propose two prompt-based approaches to selecting high-quality questions from a set of LLM-generated candidates.
Our method works under the constraints of 1) a black-box (non-modifiable) question generation model and 2) lack of access to human-annotated references --- both of which are realistic limitations for real-world deployment of LLMs.
With automatic as well as human evaluations, we empirically demonstrate that our approach can effectively select questions of higher qualities than greedy generation.
\footnote{We open-source all code and annotated data on \href{https://github.com/xingdi-eric-yuan/prompt_based_quesiton_selection}{github}.}
\end{abstract}

\section{Introduction \& Related Work}

Large Language Models (LLMs) have recently gained tremendous popularity in the NLP community  \citep{devlin2019bert, Liu2019RoBERTaAR, unilmv2, brown2020language}.
The ever-increasing size in both models and training data renders many traditional learning methods impractical/intractable. As a result, prompt-based learning has emerged as a new paradigm tailored specifically towards leveraging the power of LLMs  \citep{radford2019language, petroni2019language, raffel2020exploring, brown2020language, schick2021just, gao2021making, liu2021pre}. 
In the zero-shot setting (such as in this study), a data sample is first ``verbalized'' into an input prompt and a ground-truth response --- both often in natural language forms.
The prompt is then issued to a pre-trained LLM to obtain a predicted response, which can then be compared to the ground-truth for evaluation.
This new technique has been successfully applied to many applications including text classification \citep{yin2019benchmarking, schick2021exploiting}, QA \citep{jiang2021know}, natural language generation \citep{li2021prefix} and NLG evaluation \citep{yuan2021bartscore}.



Despite the impressive results on popular NLP benchmarks, however, the back-end LLMs are usually pre-trained with general-domain data, leading to sub-optimal performance in new domains for prompt-based learning.
There are two major challenges in successful domain adaptation.
Firstly, aside from the many known issues of LLMs  \citep{webson2021prompt,min2022rethinking,zhao2021calibrate, lampinen2022can}, their sheer size and/or accessibility (e.g., served via API over the internet) makes it prohibitively expensive and impractical for domain adaptation.
These limitations have inspired a recent line of work known as prompt editing/tuning \citep{gao2021making,li2021prefix,madaan2022memory}. 
The general idea is to systematically study the correlation between prompt construction and the performance on a specific task.
Prompt construction comes in a wide variety of flavours ranging from adapting real-valued \emph{prompt embeddings} to the order/wording/etc.\ of few-shot in-context learning examples.
Meanwhile, it also introduces a second challenge: prompt-tuning often relies on the availability of ground-truth labels of the data, which imposes much uncertainty in applications where labeled data are scarce.


\rev{
Given the ubiquity of the aforementioned challenges, we focus our study on alleviating the constraints on both annotation availability and access to model parameters, and consequently making LLMs more accessible to be deployed and used in real-world applications.
We take a mainstream NLG task, namely question generation, as a case study \citep{du2017learning,yuan2017machine,du2018harvesting,pan2019recent,liu2020asking,pyatkin2021asking}.
In this task, a model is trained to generate a natural language question conditioned on a context and an answer, such that the generated question can be answered by the provided answer using the context as supporting evidence.
Question generation is the corner stone for many NLP applications including education \cite{kurdi2020systematic,abdelghani2022conversational}, automatically FAQ generation \citep{mass2020unsupervised}, information seeking \citep{qi2020stay}, etc.
In an educational setting, for example, a question generation system can generate demonstrations that inspire students' curiosity and thinking (teaching), or to help assess students' proficiency on certain knowledge or skills (examining).
These use cases would benefit greatly from reduced dependency on computing resources, data availability, and the required expertise for fine-tuning an LM.
}

\rev{
To align with these real-world scenarios, our goal is to obtain better outputs from an inference-only LLM (i.e., as a ``black-box'', which is relatively more accessible, e.g., through online APIs).
In particular, given the common practice of sampling multiple outputs for improved generation diversity,
we propose a method that aims at selecting the best candidate based on multiple aspects of question quality in a zero-shot manner  --- notably without model adaptation or human annotations.
Our method can be seen as a post-hoc selection process within a larger NLG pipeline, and thus is orthogonal and applicable to zero-shot and in-context learning methods \citep{rubin2021learning, lu2022fantastically, liu2022makes}.
}


\section{Problem Setting}
\label{sec:problem}

\paragraph{Notations}
Formally, we consider a dataset of context-answer pairs $(c, a)$ both as strings.
The task of question generation is to generate a question $q$ that can be answered by $a$ using $c$ as supporting evidence. 
We use an off-the-shelf pre-trained LLM-based question generator in a zero-shot setting (prompt construction detailed in Appendix~\ref{sec:app:prompt}).
To simulate the black-box generator scenario, we refrain from any form of model tuning.
We do, however, assume access to a set of output sequences stochastically sampled from the question generator.
We thus ground our study to this application scenario by sampling $k$ questions $Q = \{q_i:i=1,\dots,k\}$.
For comparison as a baseline, we also denote $q_g$ as the question generated with a greedy algorithm (i.e., generating the most probable token at each time step).

Our goal is to devise an algorithm $S$ which selects the best candidate $q_{i^*}$ that maximizes some evaluation metric $M:Q\mapsto\mathbb{R}$, i.e., $S(Q)=i^*=\argmax_iM(q_i)$.
We use \Msmean, \Msmax, and \Msmin to denote the mean, min, and max of $\{M(q):q\in Q\}$, resp., and \Mgrdy for the greedy output $M(q_g)$. Semantically, \Msmin$\leq$\Msmean$\leq$\Msmax is tautologically true, and a positive result on the design of $S$ would translate to $M(q_{S(Q)})$ outperforming both \Msmean and \Mgrdy.



\paragraph{Datasets}
In this work, we adopt two question generation datasets with distinctive characteristics, namely \squad \citep{rajpurkar2016squad} and \ftqa \citep{xu2022fantastic}.
\squad was originally proposed as an extractive question answering (QA) dataset.
In the question generation literature \citep{du2018harvesting,yuan2017machine,unilmv2}, it has been used as a \textit{sentence-level} question generation task, i.e., a context $c$ is a single sentence that contains the corresponding answer $a$ as a sub-string.
\ftqa has also been used for both question answering and question generation.
It features \textit{paragraph-level question generation} (with $c$ being one or more paragraphs), and the answer $a$ is not necessarily a sub-string of $c$.
Since we do not perform any form of model/prompt tuning, we use the testing split for datasets, which consist of 11,877 data points for \squad and 1,007 for \ftqa.

\paragraph{Model}
We leverage a pre-trained \gpt model \citep{brown2020language} for both question generation and selection (detailed in \S\ref{sec:method}).
In all our experiments, we prompt the \gpt model in a 0-shot manner.
Details on all our prompts are provided in Appendix~\ref{sec:app:prompt}.

\paragraph{Evaluation Metrics \label{sec:problem:Metrics}}
We use two quantitative methods to evaluate the selected question $q'=M(q_{S(Q)})$:\\
$\bullet$ Reference-based evaluation:
Following prior works, we use \bleu for \squad \cite{du2018harvesting,unilmv2} and \rougel for \ftqa \citep{xu2022fantastic}. 
These metrics compare $q'$ against the reference question $\hat{q}$ (a.k.a. the ``ground-truth'' question in the existing literature).\\
$\bullet$ Human evaluation: we solicit human annotations on a subset of the data.
\rev{We postulate that an over-all score given holistically to rate a question would be highly subjective and thus less likely to induce agreement among annotators.}
Accordingly, we decompose the quality of a question into seven dimensions\footnote{Namely, grammatical correctness, offensiveness, clarity, relevance, importance, specificity, and answerability.}, and ask human annotators to rate a question on each dimension followed by an overall rating of the question.
We collect three annotations from different annotators for each data points.
We provide details of the human study in Appendix~\ref{sec:app:human_study}.

\begin{figure}[!t]
    \centering
    \includegraphics[width=0.5\textwidth]{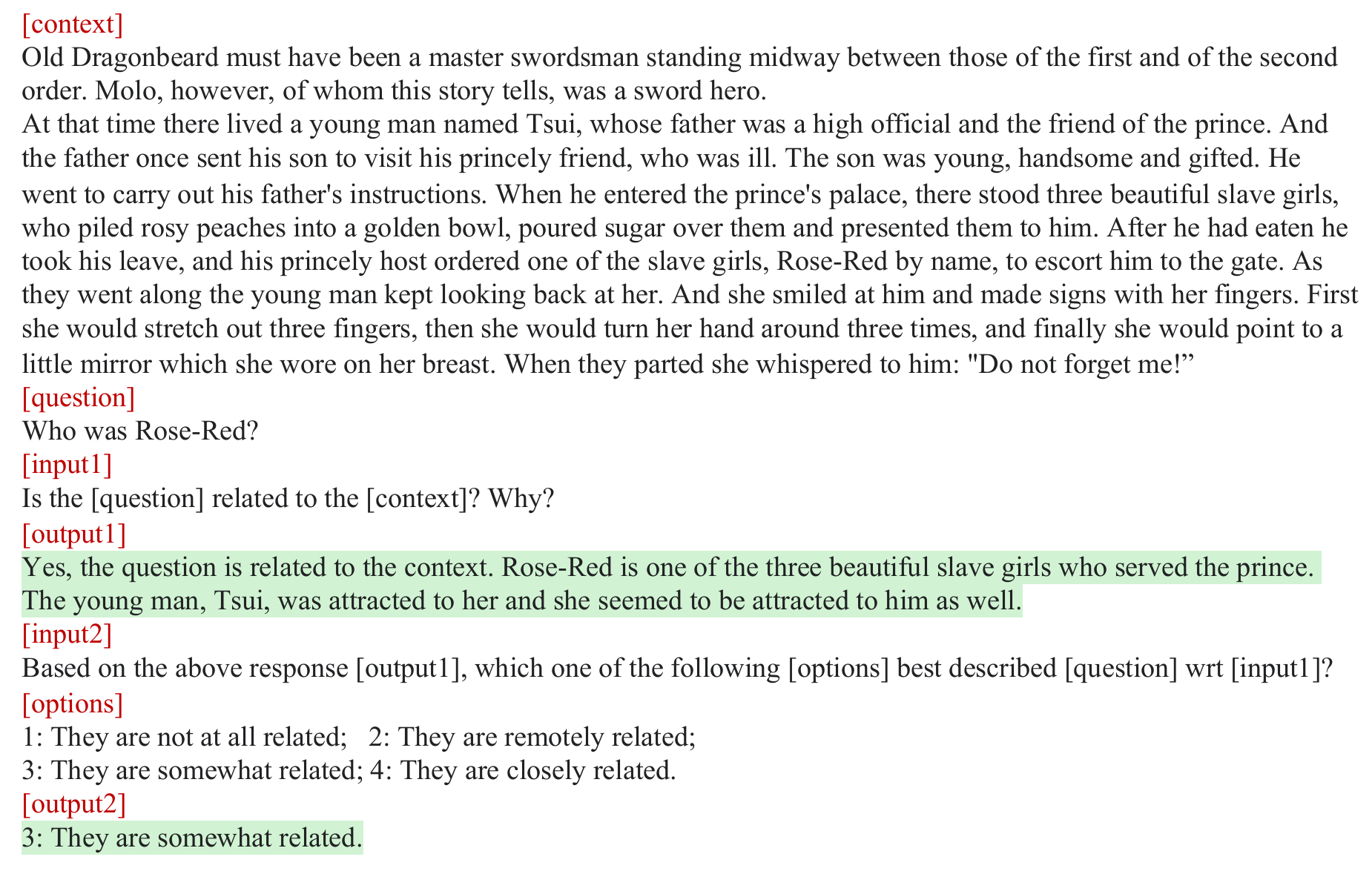}
    \caption{Prompting \gpt to rate a question's relevance. \gpt output is highlighted in green.}
    \label{fig:prompt_score}
    \vspace{-0.3em}
\end{figure}

\section{Method}
\label{sec:method}
In this section we propose three question selection methods. 
As described in \S\ref{sec:problem}, each method is used to score $k$ sampled questions in $Q$ and the candidate with the highest score is proposed as the final output.

\paragraph{\ngram similarity}
We use \ngram similarity between a question and its corresponding context to measure their relevance.
This method reflects the intuitive assumption that favorable question be closely related to the information provided by the context.
Specifically, we extract all unique $n$-grams\footnote{In all our experiments $n$ ranges from 1 to 5.} $s^n(c)$ from a given context $c$, $s^n(q)$ from a question $q$.
The \ngram similarity score is then defined as:
\begin{equation}
\small
    \text{sim}^n = \frac{|s^n(c) \cap s^n(q)|}{|s^n(q)|},
\end{equation}
where $|s|$ indicates the size of set $s$.

\begin{table}[t!]
    \centering
    \scriptsize
    \begin{tabular}{l|cc}
        \toprule
         & \squad & \ftqa \\
         & (\bleu) & (\rougel) \\
        \midrule
        \multicolumn{3}{c}{prior works (models trained/fine-tuned on these datasets)}\\
        \midrule
        \cite{du2018harvesting} & 0.152 &  -- \\
        \cite{zhang2019addressing} & 0.184 & -- \\
        UniLM Large \cite{unilmv2} & 0.228 &  -- \\
        UniLM v2 Base \cite{unilmv2} &  0.244 & -- \\
        ERNIE-GEN Large \cite{xiao2021ernie_gen} & 0.254 & -- \\
        BART \cite{xu2022fantastic} & -- & 0.527 \\
        \midrule
        \multicolumn{3}{c}{baselines (notations defined in \S\ref{sec:problem})}\\
        \midrule
        \Mgrdy (greedy) & 0.372 & 0.424\\
        \Msmean (sample avg) & 0.359 & 0.399\\
        \Msmin (lowerbound) & 0.225 & 0.259 \\
        \Msmax (upperbound) & 0.496 & 0.548 \\
        \midrule
        \multicolumn{3}{c}{question selection}\\
        \midrule
        bi-gram & 0.382 & 0.403 \\
        tri-gram & 0.380 & 0.403 \\
        round-trip & 0.392 & 0.434 \\
        overall prompt score (OPS) & 0.373 & 0.399 \\
        averaged prompt score (APS) & 0.380 & 0.406 \\
        \midrule
        \multicolumn{3}{c}{ensemble multiple methods}\\
        \midrule
        APS + round-trip& 0.397 & \textbf{0.439} \\
        bi-gram + round-trip & \underline{0.400} & 0.429\\
        tri-gram + round-trip & 0.398 & 0.430\\
        bi-gram + APS & 0.384 & 0.406\\
        tri-gram + APS & 0.383 & 0.409\\
        bi-gram + APS + round-trip & \textbf{0.401} & 0.431 \\
        tri-gram + APS + round-trip & \underline{0.400} & \underline{0.435} \\
        \bottomrule
    \end{tabular}
    \caption{Reference-based evaluation scores. Best and second best numbers (excluding baselines) are highlighted with \textbf{boldface} and \underline{underline}, respectively.}
    \label{tab:ref_based_eval}
    \vspace{-2em}
\end{table}

\paragraph{Round-trip}
Intuitively, the answer to a generated question should be semantically equivalent to the answer that has been used to generated the question.
Formally, a question generation model $\text{QG}$ and a $\text{QA}$ model (both with reasonable performance) should satisfy the following:
\begin{equation}
\small
    q' = \text{QG}(c, a); \quad
    a' = \text{QA}(c, q'); \quad
    a' = a.
\end{equation}
This idea is closely related to \textit{cycle consistency} in the existing literature on image generation \citep{zhu2017unpaired}, machine translation \citep{artetxe2018unsupervised}, and QA \citep{alberti2019synthetic,shah2019cycle}).
Here, we use \gpt as an off-the-shelf QA model to obtain $a'$ for each pair of $c$ and $q'$,
resulting in $k$ answers $A=\{a'_1,\dots,a'_k\}$ for the $k$ sampled questions in $Q$.
We then measure the similarity between each $a'_i$ and the ground-truth answer $a$ using $F_1$ score for \squad and \rougel for \ftqa (in accordance with the evaluation setup from the original papers for the two datasets).
Finally, we select the question corresponding to the generated answer $a'_{i^*}$ that overlaps the most with $a$ (i.e., that can be best answered by \gpt)
Prompts used in these experiments are detailed in Appendix~\ref{sec:app:prompt}.

\begin{figure*}[!t]
    \centering
    \includegraphics[width=0.9\textwidth]{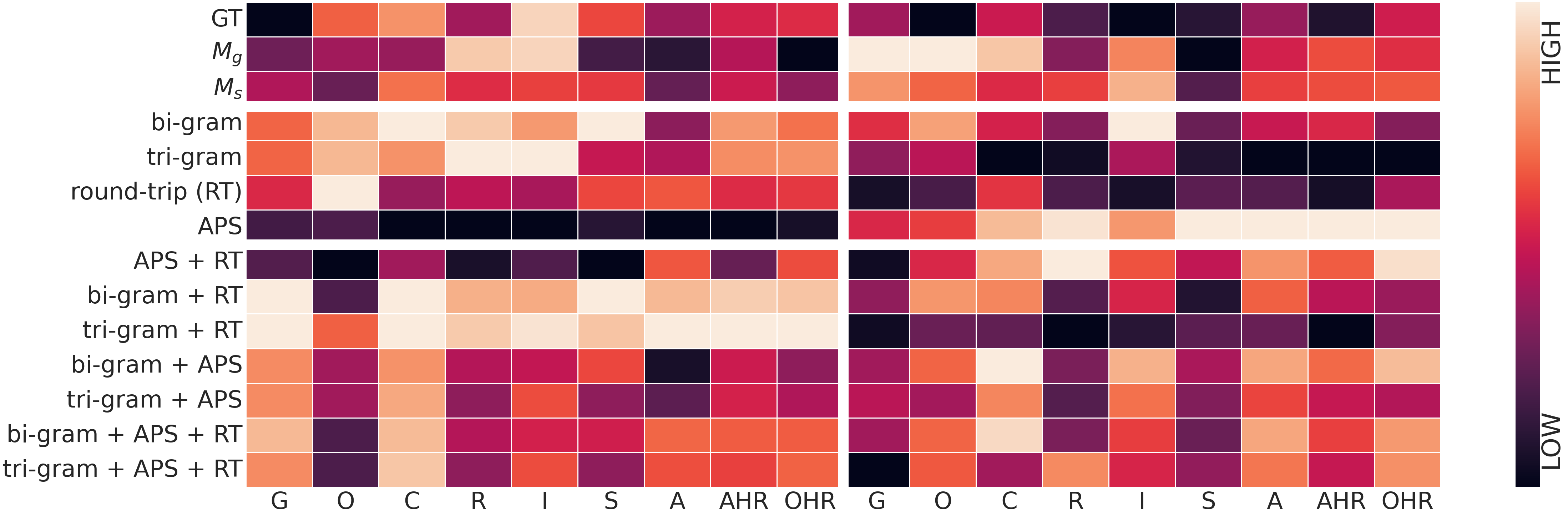}
    \caption{Human evaluation results, averaged over three annotators' scores, normalized per column. Left: \squad; right: \ftqa.  Abbreviations in x-axis denote \textbf{G}rammatical correctness, \textbf{O}ffensiveness, \textbf{C}larity, \textbf{R}elevance, \textbf{I}mportance, \textbf{S}pecificity, \textbf{A}nswerability, \textbf{A}veraged \textbf{H}uman \textbf{R}ating (over all dimensions to the left), \textbf{O}verall \textbf{H}uman \textbf{R}ating (an overall score given by annotators). Exact scores are provided in Appendix~\ref{sec:app:additional_res}.}
    \label{fig:human_eval}
    \vspace{-0.3em}
\end{figure*}

\paragraph{Prompt-based Score}


We propose a two-step procedure (Figure~\ref{fig:prompt_score}) for prompting \gpt to answer the same set of meta-questions (i.e., questions about the quality of a given question) used for human evaluation (\S\ref{sec:problem}).

In step 1, given a context-question pair, \gpt is prompted to answer a meta-question as an open question (as opposed to choosing among a list of options) as well as to verbalize a reason for its answer. 
In step 2, \gpt is prompted to choose from a list of options representing the rating scale of the meta-question.

We empirically observe that without the first step, \gpt output tends to have a low-entropy distribution, i.e., often choosing the same option for a given meta-question disregarding the different context-question pairs.
In contrast, the model appears to be better primed wrt output diversity with the additional first-step, which is inline with observations made in some existing studies \citep{nye2021show,wei2022chain}.


Similar to human evaluation, we also prompt \gpt to generate an overall score to a question.
We use \textit{overall prompt-based score (OPS)} to denote this \gpt-labeled score, and \textit{averaged prompt-based score (APS)} to denote the average score over all individual meta-questions.

\section{Results and Discussion}
\label{sec:results}

To measure the performance of a selection method (\S\ref{sec:method}), we use it to select one out of $k$ questions stochastically sampled from \gpt, and score the selection with the evaluation metrics outlined in \S\ref{sec:problem:Metrics}. We set $k=5$ for all our experiments.
\rev{Additionally, we test the ensemble performance with multiple methods. 
To ensure comparability, we normalize the scores obtained from each selection method into the range between 0 and 1, and use their average score to perform question selection.}

\subsection{Reference-based evaluation}
Reference-based evaluation are automatic metrics that are applied to the entire test sets of \squad and \ftqa.
We observe in Table~\ref{tab:ref_based_eval} that on both datasets, all question selection methods outperform \Msmean, \rev{the average score over all five sampled questions,} validating the effectiveness of the proposed methods.
While all individual methods outperform \rev{the greedy generation baseline} \Mgrdy on \squad, round-trip is the best performing one, outperforming \Mgrdy on both datasets.
It can be further improved via ensemble with \ngram and/or prompt-based scores (using uniform weights).

Note that prior studies require a large amount of labeled data for model training/fine-tuning, while \gpt performs zero-shot inference. Despite this major difference in learning paradigm, most \gpt-based models proposed here outperform previous results by significant margins on the \squad dataset --- even the least performant samples \Msmin \rev{(lowerbound)} achieve competitive results.
For \ftqa, however, only the best samples \Msmax \rev{(upperbound)} outperform previous results \citep{xu2022fantastic}, indicating margins for improvement on question selection strategies for future work.

\subsection{Human Evaluation}
Human evaluation consists of $16,800$ annotations (collected from 87 annotators) evenly split across the two datasets (details in Appendix~\ref{sec:app:human_study}).
For question generation (among many language generation tasks), model outputs may exhibit linguistic diversity while maintaining semantic equivalence.
It is thus highly problematic to evaluate such outputs against a single reference (i.e, ``ground-truth'' questions).
Figure~\ref{fig:human_eval} empirically shows that the ground-truth (GT) questions provided in the datasets often fail to receive the highest human ratings,
on many occasions scoring lower than stochastic samples from \gpt (\Msmean).
Consequently, we strongly advocate for human evaluation, which we believe is higly effective in improving generalizability of our results to real-world applications.

Another prominent observation is that \ngram and APS perform quite differently on the two datasets.
On \squad, \ngram similarity outperforms other individual methods,
with further noticeable improvements via ensemble with round-trip. 
APS, on the other hand, does not work nearly as well, performing the worst for almost all meta-questions.
In contrast, \ngram (particularly tri-gram) similarity shows the worst performance on \ftqa, while APS outperforms all other methods by a noticeable margin.

We posit that the reversed trend in comparing \ngram and APS can be explained by the distinct natures of the datasets.
For \squad, the sentence-level contexts are relatively short and simple with strictly extractive answers (i.e., the answers being sub-strings of the corresponding contexts).
As a result, paraphrasing the context can be a rather effective question generation strategy, hence the stronger correlation between question quality and the $c$--$q$ \ngram similarity.
On the other hand, with multi-paragraph contexts and abstractive, open-ended answers, 
questions are more likely posed about abstract ideas rather than simple context paraphrasing. Consequently, \ngram similarity, which favors local context paraphrasing, can no longer serve as a good question selection strategy. 

\subsection{Limitations and Future Work}
We acknowledge that our system has some limitations that warrants further investigation.
For example, one needs to be mindful of the specific downstream applications of the proposed methods, both in terms of potentially large variance in out-of-distribution performance (e.g. \textit{divergent} question generation, \citealt{abdelghani2022conversational}) and of mitigating harmful/toxic contents in educational applications \citep{bender2021dangers}.

We also acknowledge the prohibitively restrictive access to the GPT-3 model at the time of writing. We do believe that this constraint will relax over time, and meanwhile, hoping that our proposal can shed light on research and applications with more accessible LLMs such as GPT-J \citep{gpt-j} and BLOOM \citep{bloom} for future work.



\rev{
\section{Conclusion}
In this study, we investigate the practical problem of selecting the best output from multiple samples generated by an LLM. 
Using question generation as a case study, we propose two prompt-based approaches that select high-quality questions according to question quality from multiple perspectives.
To alleviate real-world constraints on using large LMs such as computation resources and data availability, the proposed methods do not rely on model fine-tuning nor human annotation.
Extensive experiments with both automatic and human evaluations evince the effectiveness of our approach on selecting high-quality questions from stochastic samples.
}

\bibliography{custom}

\begin{thebibliography}{46}
\expandafter\ifx\csname natexlab\endcsname\relax\def\natexlab#1{#1}\fi

\bibitem[{Abdelghani et~al.(2022)Abdelghani, Oudeyer, Law, de~Vulpillieres, and
  Sauz{\'e}on}]{abdelghani2022conversational}
Rania Abdelghani, Pierre-Yves Oudeyer, Edith Law, Catherine de~Vulpillieres,
  and H{\'e}lene Sauz{\'e}on. 2022.
\newblock Conversational agents for fostering curiosity-driven learning in
  children.
\newblock \emph{International Journal of Human-Computer Studies}.

\bibitem[{Alberti et~al.(2019)Alberti, Andor, Pitler, Devlin, and
  Collins}]{alberti2019synthetic}
Chris Alberti, Daniel Andor, Emily Pitler, Jacob Devlin, and Michael Collins.
  2019.
\newblock \href {https://doi.org/10.18653/v1/P19-1620} {Synthetic {QA} corpora
  generation with roundtrip consistency}.
\newblock In \emph{Proceedings of the 57th Annual Meeting of the Association
  for Computational Linguistics}, pages 6168--6173, Florence, Italy.
  Association for Computational Linguistics.

\bibitem[{Artetxe et~al.(2018)Artetxe, Labaka, Agirre, and
  Cho}]{artetxe2018unsupervised}
Mikel Artetxe, Gorka Labaka, Eneko Agirre, and Kyunghyun Cho. 2018.
\newblock \href {https://openreview.net/forum?id=Sy2ogebAW} {Unsupervised
  neural machine translation}.
\newblock In \emph{International Conference on Learning Representations}.

\bibitem[{Bao et~al.(2020)Bao, Dong, Wei, Wang, Yang, Liu, Wang, Gao, Piao,
  Zhou, and Hon}]{unilmv2}
Hangbo Bao, Li~Dong, Furu Wei, Wenhui Wang, Nan Yang, Xiaodong Liu, Yu~Wang,
  Jianfeng Gao, Songhao Piao, Ming Zhou, and Hsiao-Wuen Hon. 2020.
\newblock \href {https://proceedings.mlr.press/v119/bao20a.html} {{U}ni{LM}v2:
  Pseudo-masked language models for unified language model pre-training}.
\newblock In \emph{Proceedings of the 37th International Conference on Machine
  Learning}, volume 119 of \emph{Proceedings of Machine Learning Research},
  pages 642--652. PMLR.

\bibitem[{Bender et~al.(2021)Bender, Gebru, McMillan-Major, and
  Shmitchell}]{bender2021dangers}
Emily~M Bender, Timnit Gebru, Angelina McMillan-Major, and Shmargaret
  Shmitchell. 2021.
\newblock On the dangers of stochastic parrots: Can language models be too big?
\newblock In \emph{Proceedings of the 2021 ACM Conference on Fairness,
  Accountability, and Transparency}, pages 610--623.

\bibitem[{BigScience(2022)}]{bloom}
BigScience. 2022.
\newblock Bigscience language open-science open-access multilingual (bloom)
  language model.
\newblock \emph{International}.

\bibitem[{Brown et~al.(2020)Brown, Mann, Ryder, Subbiah, Kaplan, Dhariwal,
  Neelakantan, Shyam, Sastry, Askell et~al.}]{brown2020language}
Tom Brown, Benjamin Mann, Nick Ryder, Melanie Subbiah, Jared~D Kaplan, Prafulla
  Dhariwal, Arvind Neelakantan, Pranav Shyam, Girish Sastry, Amanda Askell,
  et~al. 2020.
\newblock Language models are few-shot learners.
\newblock \emph{Advances in neural information processing systems},
  33:1877--1901.

\bibitem[{Devlin et~al.(2019)Devlin, Chang, Lee, and
  Toutanova}]{devlin2019bert}
Jacob Devlin, Ming-Wei Chang, Kenton Lee, and Kristina Toutanova. 2019.
\newblock \href {https://doi.org/10.18653/v1/N19-1423} {{BERT}: Pre-training of
  deep bidirectional transformers for language understanding}.
\newblock In \emph{Proceedings of the 2019 Conference of the North {A}merican
  Chapter of the Association for Computational Linguistics: Human Language
  Technologies, Volume 1 (Long and Short Papers)}, pages 4171--4186,
  Minneapolis, Minnesota. Association for Computational Linguistics.

\bibitem[{Du and Cardie(2018)}]{du2018harvesting}
Xinya Du and Claire Cardie. 2018.
\newblock \href {https://doi.org/10.18653/v1/P18-1177} {Harvesting
  paragraph-level question-answer pairs from {W}ikipedia}.
\newblock In \emph{Proceedings of the 56th Annual Meeting of the Association
  for Computational Linguistics (Volume 1: Long Papers)}, pages 1907--1917,
  Melbourne, Australia. Association for Computational Linguistics.

\bibitem[{Du et~al.(2017)Du, Shao, and Cardie}]{du2017learning}
Xinya Du, Junru Shao, and Claire Cardie. 2017.
\newblock Learning to ask: Neural question generation for reading
  comprehension.
\newblock \emph{arXiv preprint arXiv:1705.00106}.

\bibitem[{Gao et~al.(2021)Gao, Fisch, and Chen}]{gao2021making}
Tianyu Gao, Adam Fisch, and Danqi Chen. 2021.
\newblock \href {https://doi.org/10.18653/v1/2021.acl-long.295} {Making
  pre-trained language models better few-shot learners}.
\newblock In \emph{Proceedings of the 59th Annual Meeting of the Association
  for Computational Linguistics and the 11th International Joint Conference on
  Natural Language Processing (Volume 1: Long Papers)}, pages 3816--3830,
  Online. Association for Computational Linguistics.

\bibitem[{Jiang et~al.(2021)Jiang, Araki, Ding, and Neubig}]{jiang2021know}
Zhengbao Jiang, Jun Araki, Haibo Ding, and Graham Neubig. 2021.
\newblock \href {https://doi.org/10.1162/tacl_a_00407} {How can we know when
  language models know? on the calibration of language models for question
  answering}.
\newblock \emph{Transactions of the Association for Computational Linguistics},
  9:962--977.

\bibitem[{Kurdi et~al.(2020)Kurdi, Leo, Parsia, Sattler, and
  Al-Emari}]{kurdi2020systematic}
Ghader Kurdi, Jared Leo, Bijan Parsia, Uli Sattler, and Salam Al-Emari. 2020.
\newblock A systematic review of automatic question generation for educational
  purposes.
\newblock \emph{International Journal of Artificial Intelligence in Education},
  30(1):121--204.

\bibitem[{Lampinen et~al.(2022)Lampinen, Dasgupta, Chan, Matthewson, Tessler,
  Creswell, McClelland, Wang, and Hill}]{lampinen2022can}
Andrew~K Lampinen, Ishita Dasgupta, Stephanie~CY Chan, Kory Matthewson,
  Michael~Henry Tessler, Antonia Creswell, James~L McClelland, Jane~X Wang, and
  Felix Hill. 2022.
\newblock Can language models learn from explanations in context?
\newblock \emph{arXiv preprint arXiv:2204.02329}.

\bibitem[{Li and Liang(2021)}]{li2021prefix}
Xiang~Lisa Li and Percy Liang. 2021.
\newblock \href {https://doi.org/10.18653/v1/2021.acl-long.353} {Prefix-tuning:
  Optimizing continuous prompts for generation}.
\newblock In \emph{Proceedings of the 59th Annual Meeting of the Association
  for Computational Linguistics and the 11th International Joint Conference on
  Natural Language Processing (Volume 1: Long Papers)}, pages 4582--4597,
  Online. Association for Computational Linguistics.

\bibitem[{Liu et~al.(2020)Liu, Wei, Niu, Chen, and He}]{liu2020asking}
Bang Liu, Haojie Wei, Di~Niu, Haolan Chen, and Yancheng He. 2020.
\newblock Asking questions the human way: Scalable question-answer generation
  from text corpus.
\newblock In \emph{Proceedings of The Web Conference 2020}, pages 2032--2043.

\bibitem[{Liu et~al.(2022)Liu, Shen, Zhang, Dolan, Carin, and
  Chen}]{liu2022makes}
Jiachang Liu, Dinghan Shen, Yizhe Zhang, Bill Dolan, Lawrence Carin, and Weizhu
  Chen. 2022.
\newblock \href {https://doi.org/10.18653/v1/2022.deelio-1.10} {What makes good
  in-context examples for {GPT}-3?}
\newblock In \emph{Proceedings of Deep Learning Inside Out (DeeLIO 2022): The
  3rd Workshop on Knowledge Extraction and Integration for Deep Learning
  Architectures}, pages 100--114, Dublin, Ireland and Online. Association for
  Computational Linguistics.

\bibitem[{Liu et~al.(2021)Liu, Yuan, Fu, Jiang, Hayashi, and
  Neubig}]{liu2021pre}
Pengfei Liu, Weizhe Yuan, Jinlan Fu, Zhengbao Jiang, Hiroaki Hayashi, and
  Graham Neubig. 2021.
\newblock Pre-train, prompt, and predict: A systematic survey of prompting
  methods in natural language processing.
\newblock \emph{arXiv preprint arXiv:2107.13586}.

\bibitem[{Liu et~al.(2019)Liu, Ott, Goyal, Du, Joshi, Chen, Levy, Lewis,
  Zettlemoyer, and Stoyanov}]{Liu2019RoBERTaAR}
Yinhan Liu, Myle Ott, Naman Goyal, Jingfei Du, Mandar Joshi, Danqi Chen, Omer
  Levy, Mike Lewis, Luke Zettlemoyer, and Veselin Stoyanov. 2019.
\newblock Roberta: A robustly optimized bert pretraining approach.
\newblock \emph{ArXiv}, abs/1907.11692.

\bibitem[{Lu et~al.(2022)Lu, Bartolo, Moore, Riedel, and
  Stenetorp}]{lu2022fantastically}
Yao Lu, Max Bartolo, Alastair Moore, Sebastian Riedel, and Pontus Stenetorp.
  2022.
\newblock \href {https://doi.org/10.18653/v1/2022.acl-long.556} {Fantastically
  ordered prompts and where to find them: Overcoming few-shot prompt order
  sensitivity}.
\newblock In \emph{Proceedings of the 60th Annual Meeting of the Association
  for Computational Linguistics (Volume 1: Long Papers)}, pages 8086--8098,
  Dublin, Ireland. Association for Computational Linguistics.

\bibitem[{Madaan et~al.(2022)Madaan, Tandon, Clark, and
  Yang}]{madaan2022memory}
Aman Madaan, Niket Tandon, Peter Clark, and Yiming Yang. 2022.
\newblock Memory-assisted prompt editing to improve gpt-3 after deployment.
\newblock \emph{arXiv preprint arXiv:2201.06009}.

\bibitem[{Mass et~al.(2020)Mass, Carmeli, Roitman, and
  Konopnicki}]{mass2020unsupervised}
Yosi Mass, Boaz Carmeli, Haggai Roitman, and David Konopnicki. 2020.
\newblock Unsupervised faq retrieval with question generation and bert.
\newblock In \emph{Proceedings of the 58th Annual Meeting of the Association
  for Computational Linguistics}, pages 807--812.

\bibitem[{Min et~al.(2022)Min, Lyu, Holtzman, Artetxe, Lewis, Hajishirzi, and
  Zettlemoyer}]{min2022rethinking}
Sewon Min, Xinxi Lyu, Ari Holtzman, Mikel Artetxe, Mike Lewis, Hannaneh
  Hajishirzi, and Luke Zettlemoyer. 2022.
\newblock Rethinking the role of demonstrations: What makes in-context learning
  work?
\newblock \emph{arXiv preprint arXiv:2202.12837}.

\bibitem[{Nye et~al.(2021)Nye, Andreassen, Gur-Ari, Michalewski, Austin,
  Bieber, Dohan, Lewkowycz, Bosma, Luan et~al.}]{nye2021show}
Maxwell Nye, Anders~Johan Andreassen, Guy Gur-Ari, Henryk Michalewski, Jacob
  Austin, David Bieber, David Dohan, Aitor Lewkowycz, Maarten Bosma, David
  Luan, et~al. 2021.
\newblock Show your work: Scratchpads for intermediate computation with
  language models.
\newblock \emph{arXiv preprint arXiv:2112.00114}.

\bibitem[{Pan et~al.(2019)Pan, Lei, Chua, and Kan}]{pan2019recent}
Liangming Pan, Wenqiang Lei, Tat-Seng Chua, and Min-Yen Kan. 2019.
\newblock Recent advances in neural question generation.
\newblock \emph{arXiv preprint arXiv:1905.08949}.

\bibitem[{Petroni et~al.(2019)Petroni, Rockt{\"a}schel, Riedel, Lewis, Bakhtin,
  Wu, and Miller}]{petroni2019language}
Fabio Petroni, Tim Rockt{\"a}schel, Sebastian Riedel, Patrick Lewis, Anton
  Bakhtin, Yuxiang Wu, and Alexander Miller. 2019.
\newblock \href {https://doi.org/10.18653/v1/D19-1250} {Language models as
  knowledge bases?}
\newblock In \emph{Proceedings of the 2019 Conference on Empirical Methods in
  Natural Language Processing and the 9th International Joint Conference on
  Natural Language Processing (EMNLP-IJCNLP)}, pages 2463--2473, Hong Kong,
  China. Association for Computational Linguistics.

\bibitem[{Pyatkin et~al.(2021)Pyatkin, Roit, Michael, Tsarfaty, Goldberg, and
  Dagan}]{pyatkin2021asking}
Valentina Pyatkin, Paul Roit, Julian Michael, Reut Tsarfaty, Yoav Goldberg, and
  Ido Dagan. 2021.
\newblock Asking it all: Generating contextualized questions for any semantic
  role.
\newblock \emph{arXiv preprint arXiv:2109.04832}.

\bibitem[{Qi et~al.(2020)Qi, Zhang, and Manning}]{qi2020stay}
Peng Qi, Yuhao Zhang, and Christopher~D. Manning. 2020.
\newblock \href {https://doi.org/10.18653/v1/2020.findings-emnlp.3} {Stay
  hungry, stay focused: Generating informative and specific questions in
  information-seeking conversations}.
\newblock In \emph{Findings of the Association for Computational Linguistics:
  EMNLP 2020}, pages 25--40, Online. Association for Computational Linguistics.

\bibitem[{Radford et~al.(2019)Radford, Wu, Child, Luan, Amodei, Sutskever
  et~al.}]{radford2019language}
Alec Radford, Jeffrey Wu, Rewon Child, David Luan, Dario Amodei, Ilya
  Sutskever, et~al. 2019.
\newblock Language models are unsupervised multitask learners.
\newblock \emph{OpenAI blog}, 1(8):9.

\bibitem[{Raffel et~al.(2020)Raffel, Shazeer, Roberts, Lee, Narang, Matena,
  Zhou, Li, and Liu}]{raffel2020exploring}
Colin Raffel, Noam Shazeer, Adam Roberts, Katherine Lee, Sharan Narang, Michael
  Matena, Yanqi Zhou, Wei Li, and Peter~J. Liu. 2020.
\newblock \href {http://jmlr.org/papers/v21/20-074.html} {Exploring the limits
  of transfer learning with a unified text-to-text transformer}.
\newblock \emph{Journal of Machine Learning Research}, 21(140):1--67.

\bibitem[{Rajpurkar et~al.(2016)Rajpurkar, Zhang, Lopyrev, and
  Liang}]{rajpurkar2016squad}
Pranav Rajpurkar, Jian Zhang, Konstantin Lopyrev, and Percy Liang. 2016.
\newblock \href {https://doi.org/10.18653/v1/D16-1264} {{SQ}u{AD}: 100,000+
  questions for machine comprehension of text}.
\newblock In \emph{Proceedings of the 2016 Conference on Empirical Methods in
  Natural Language Processing}, pages 2383--2392, Austin, Texas. Association
  for Computational Linguistics.

\bibitem[{Rubin et~al.(2021)Rubin, Herzig, and Berant}]{rubin2021learning}
Ohad Rubin, Jonathan Herzig, and Jonathan Berant. 2021.
\newblock Learning to retrieve prompts for in-context learning.
\newblock \emph{arXiv preprint arXiv:2112.08633}.

\bibitem[{Schick and Sch{\"u}tze(2021{\natexlab{a}})}]{schick2021exploiting}
Timo Schick and Hinrich Sch{\"u}tze. 2021{\natexlab{a}}.
\newblock \href {https://doi.org/10.18653/v1/2021.eacl-main.20} {Exploiting
  cloze-questions for few-shot text classification and natural language
  inference}.
\newblock In \emph{Proceedings of the 16th Conference of the European Chapter
  of the Association for Computational Linguistics: Main Volume}, pages
  255--269, Online. Association for Computational Linguistics.

\bibitem[{Schick and Sch{\"u}tze(2021{\natexlab{b}})}]{schick2021just}
Timo Schick and Hinrich Sch{\"u}tze. 2021{\natexlab{b}}.
\newblock \href {https://doi.org/10.18653/v1/2021.naacl-main.185} {It{'}s not
  just size that matters: Small language models are also few-shot learners}.
\newblock In \emph{Proceedings of the 2021 Conference of the North American
  Chapter of the Association for Computational Linguistics: Human Language
  Technologies}, pages 2339--2352, Online. Association for Computational
  Linguistics.

\bibitem[{Shah et~al.(2019)Shah, Chen, Rohrbach, and Parikh}]{shah2019cycle}
Meet Shah, Xinlei Chen, Marcus Rohrbach, and Devi Parikh. 2019.
\newblock Cycle-consistency for robust visual question answering.
\newblock In \emph{Proceedings of the IEEE/CVF Conference on Computer Vision
  and Pattern Recognition}, pages 6649--6658.

\bibitem[{Wang and Komatsuzaki(2021)}]{gpt-j}
Ben Wang and Aran Komatsuzaki. 2021.
\newblock {GPT-J-6B: A 6 Billion Parameter Autoregressive Language Model}.
\newblock \url{https://github.com/kingoflolz/mesh-transformer-jax}.

\bibitem[{Webson and Pavlick(2021)}]{webson2021prompt}
Albert Webson and Ellie Pavlick. 2021.
\newblock Do prompt-based models really understand the meaning of their
  prompts?
\newblock \emph{arXiv preprint arXiv:2109.01247}.

\bibitem[{Wei et~al.(2022)Wei, Wang, Schuurmans, Bosma, Chi, Le, and
  Zhou}]{wei2022chain}
Jason Wei, Xuezhi Wang, Dale Schuurmans, Maarten Bosma, Ed~Chi, Quoc Le, and
  Denny Zhou. 2022.
\newblock Chain of thought prompting elicits reasoning in large language
  models.
\newblock \emph{arXiv preprint arXiv:2201.11903}.

\bibitem[{Xiao et~al.(2021)Xiao, Zhang, Li, Sun, Tian, Wu, and
  Wang}]{xiao2021ernie_gen}
Dongling Xiao, Han Zhang, Yukun Li, Yu~Sun, Hao Tian, Hua Wu, and Haifeng Wang.
  2021.
\newblock Ernie-gen: An enhanced multi-flow pre-training and fine-tuning
  framework for natural language generation.
\newblock In \emph{Proceedings of the Twenty-Ninth International Joint
  Conference on Artificial Intelligence}, IJCAI'20.

\bibitem[{Xu et~al.(2022)Xu, Wang, Yu, Ritchie, Yao, Wu, Zhang, Li, Bradford,
  Sun et~al.}]{xu2022fantastic}
Ying Xu, Dakuo Wang, Mo~Yu, Daniel Ritchie, Bingsheng Yao, Tongshuang Wu, Zheng
  Zhang, Toby Jia-Jun Li, Nora Bradford, Branda Sun, et~al. 2022.
\newblock Fantastic questions and where to find them: Fairytaleqa--an authentic
  dataset for narrative comprehension.
\newblock \emph{arXiv preprint arXiv:2203.13947}.

\bibitem[{Yin et~al.(2019)Yin, Hay, and Roth}]{yin2019benchmarking}
Wenpeng Yin, Jamaal Hay, and Dan Roth. 2019.
\newblock \href {https://doi.org/10.18653/v1/D19-1404} {Benchmarking zero-shot
  text classification: Datasets, evaluation and entailment approach}.
\newblock In \emph{Proceedings of the 2019 Conference on Empirical Methods in
  Natural Language Processing and the 9th International Joint Conference on
  Natural Language Processing (EMNLP-IJCNLP)}, pages 3914--3923, Hong Kong,
  China. Association for Computational Linguistics.

\bibitem[{Yuan et~al.(2021)Yuan, Neubig, and Liu}]{yuan2021bartscore}
Weizhe Yuan, Graham Neubig, and Pengfei Liu. 2021.
\newblock \href
  {https://proceedings.neurips.cc/paper/2021/file/e4d2b6e6fdeca3e60e0f1a62fee3d9dd-Paper.pdf}
  {Bartscore: Evaluating generated text as text generation}.
\newblock In \emph{Advances in Neural Information Processing Systems},
  volume~34, pages 27263--27277. Curran Associates, Inc.

\bibitem[{Yuan et~al.(2017)Yuan, Wang, Gulcehre, Sordoni, Bachman, Subramanian,
  Zhang, and Trischler}]{yuan2017machine}
Xingdi Yuan, Tong Wang, Caglar Gulcehre, Alessandro Sordoni, Philip Bachman,
  Sandeep Subramanian, Saizheng Zhang, and Adam Trischler. 2017.
\newblock Machine comprehension by text-to-text neural question generation.
\newblock In \emph{arXiv.}

\bibitem[{Zhang and Bansal(2019)}]{zhang2019addressing}
Shiyue Zhang and Mohit Bansal. 2019.
\newblock \href {https://doi.org/10.18653/v1/D19-1253} {Addressing semantic
  drift in question generation for semi-supervised question answering}.
\newblock In \emph{Proceedings of the 2019 Conference on Empirical Methods in
  Natural Language Processing and the 9th International Joint Conference on
  Natural Language Processing (EMNLP-IJCNLP)}, pages 2495--2509, Hong Kong,
  China. Association for Computational Linguistics.

\bibitem[{Zhao et~al.(2021)Zhao, Wallace, Feng, Klein, and
  Singh}]{zhao2021calibrate}
Zihao Zhao, Eric Wallace, Shi Feng, Dan Klein, and Sameer Singh. 2021.
\newblock Calibrate before use: Improving few-shot performance of language
  models.
\newblock In \emph{International Conference on Machine Learning}, pages
  12697--12706. PMLR.

\bibitem[{Zhu et~al.(2017)Zhu, Park, Isola, and Efros}]{zhu2017unpaired}
Jun-Yan Zhu, Taesung Park, Phillip Isola, and Alexei~A Efros. 2017.
\newblock Unpaired image-to-image translation using cycle-consistent
  adversarial networks.
\newblock In \emph{Proceedings of the IEEE international conference on computer
  vision}, pages 2223--2232.

\end{thebibliography}
\bibliographystyle{acl_natbib}

\clearpage
\appendix

\textbf{\large{Contents in Appendices:}}
\begin{itemize}
    \item In Appendix~\ref{sec:app:prompt}, we report all prompt templates we used in this work. 
    \item In Appendix~\ref{sec:app:human_study}, we provide details on the human study.
    \item In Appendix~\ref{sec:app:additional_res}, we provide the full set of our experiment results.
    \item In Appendix~\ref{sec:app:implementation}, we report implementation details.
\end{itemize}

\section{Prompt Designs}
\label{sec:app:prompt}

We report an example of our prompt for question generation in Figure~\ref{fig:prompt_qgen}.

We report an example of our prompt for QA (used in round-trip) in Figure~\ref{fig:prompt_qa}.

We report an example of our prompt in obtaining prompt scores in Figure~\ref{fig:prompt_score}.

\begin{figure*}[h]
\begin{mdframed}
    \parbox{.99\textwidth}{
    \textcolor{red1}{Story:}\\
    As soon as the lady had departed the fisher's son awoke, and the dark
    lad told him of her visit, and how he would never see her as long as he
    lived. At this the fisher's son felt the cold creeping up to his heart,
    yet he knew the fault had not been his that sleep had overtaken him.\\
    'I will search the whole world through till I find her,' cried he, and
    the dark lad laughed as he heard him. But the fisher's son took no heed,
    and off he went, following the sun day after day, till his shoes were in
    holes and his feet were sore from the journey. Nought did he see but
    the birds that made their nests in the trees, not so much as a goat or
    a rabbit. On and on and on he went, till suddenly he came upon a little
    house, with a woman standing outside it.\\
    \textcolor{red1}{Instruction:}\\
    Read the above story, ask a question and answer it.\\
    \textcolor{red1}{Question:}\\
    \colorbox{green1}{GPT-3 FILLS IN THIS BLANK}\\
    \textcolor{red1}{Answer:}\\
    search the whole world through till he found her
    }
\end{mdframed}
\caption{An example of prompting GPT-3 for question generation. We use the text before \colorbox{green1}{green} as prompt, and text after \colorbox{green1}{green} as suffix. We refer readers to \href{https://beta.openai.com/docs/api-reference/completions/create}{the GPT-3 documentation} for more details about GPT-3's  inserting completion mode.}
\label{fig:prompt_qgen}
\end{figure*}

\begin{figure*}[t!]
\begin{mdframed}
    \parbox{.99\textwidth}{
    \textcolor{red1}{[Document]:}\\
    is cheeks were red with passion, and his eyes were bright, for he could not but notice that, now that she was safe at Orphir under her true love's protection, the Lady Morna's manner had grown cold and distant again, and he was beginning to lose faith in Snorro's charm.\\
    \\
    Angry and disappointed, he had sought his mother's room to pour out his story of vexation to her.\\
    \\
    He stopped short, however, when he saw the wonderful waistcoat lying on the table, all gold and silver and shining colours. It was like a fairy garment, and its beauty took his breath away.\\
    \\
    \textcolor{red1}{[Question]:}\\
    Why did Harold lose faith in Snorro's charm?\\
    \\
    \textcolor{red1}{[Answer]:}\\
    \colorbox{green1}{Harold lost faith in Snorro's charm because the Lady Morna's manner had grown cold and distant} \\
    \colorbox{green1}{again.}
    }
\end{mdframed}
\caption{An example of prompting GPT-3 for QA. GPT-3 output is highlighted in green.}
\label{fig:prompt_qa}
\end{figure*}

\begin{figure*}[t!]
\begin{mdframed}
    \parbox{.99\textwidth}{
1. Is the question gramatically correct?\\
1) It is grammatically incorrect\\
2) It has some grammatical issues\\
3) It is grammatically correct\\
\\
2. Is the question offensive to people?\\
1) It is very offensive\\
2) It may be offensive\\
3) It is not at all offensive\\
\\
3. Is the question clear?\\
1) It is not at all clear\\
2) It is mostly clear\\
3) Is is very clear\\
\\
4. Is the question related to the context of the attached document?\\
1) It is not at all related\\
2) It is somewhat related\\
3) It is closely related\\
\\
5. Is the question asking about an important aspect of the context of the attached document?\\
1) Not at all important\\
2) It may be important\\
3) It is very important\\
\\ 
6. Is the question asking about a specific piece of information in the attached document?\\
1) The question is very generic\\
2) The question is somewhat generic\\
3) The question is very specific\\
\\
7. Can the question be answered using information in the attached document?\\
1) No, answering the question requires completely different information\\
2) The question can be partially answered using information from the document\\
3) The question can be perfectly answered using information from the document\\
\\ 
8. What is your overall rating of the question generated based on the attached document?\\
1) The question is very bad\\
2) The question is okay\\
3) The question is very good
    }
\end{mdframed}
\caption{Meta-questions we designed for human evaluation.}
\label{fig:human_eval_meta_q}
\end{figure*}

\section{Human Study}
\label{sec:app:human_study}

We randomly sample 50 documents from each of the two datasets \squad and \ftqa. Each document correspond to one ground-truth question and six questions generated by \gpt (five by stochastic sampling and one by greedy search). Each question is then rated by three human annotators wrt seven meta-questions and one over-all rating, altogether constituting $50\times 2\times (1+5+1)\times 3\times (7+1)=16,800$ annotations.
There are in total 87 annotators involved in the annotation process, all annotators are English speakers, they are recruited from regions including Europe, the United States and United Kingdom. Each annotator on average performed 193 annotations and was paid on average \$14.1 USD per hour.

We perform a basic spam filtering process on the raw annotations. 
We observe a 15.4\% spam rate. 
All human scores reported in this paper are computed after spam removal.

We report the eight meta-questions we used for human annotation in Figure~\ref{fig:human_eval_meta_q}.
The eight meta-questions correspond to columns in Figure~\ref{fig:human_eval}.
We collect three annotations from different annotators for every meta-question, we report the averaged human agreement rate in Table~\ref{tab:human_agreements}.


\begin{table}[h!]
    \centering
    \small
    \begin{tabular}{l|c}
        \toprule
        grammatical correctness & 0.698 \\
        offensiveness & 0.788 \\
        clarity & 0.640 \\
        relevance & 0.670 \\
        importance & 0.558 \\
        specificity & 0.619 \\
        answerability & 0.588 \\
        overall human rating (OHR) & 0.485 \\
        \bottomrule
    \end{tabular}
    \caption{Averaged human agreements among three annotators. An agreement indicates that all three annotators selected the same option for a meta-question.}
    \label{tab:human_agreements}
\end{table}

\section{Additional Results}
\label{sec:app:additional_res}
In Table~\ref{tab:ref_based_eval_full}, we report the full experiment results for reference-based evaluation.

In Table~\ref{tab:human_squad}, we report the full results for human evaluation on \squad.

In Table~\ref{tab:human_ft}, we report the full results for human evaluation on \ftqa.

\begin{table}[t!]
    \centering
    \scriptsize
    \begin{tabular}{l|cc}
        \toprule
         & \squad & \ftqa \\
         & (\bleu) & (\rougel) \\
        \midrule
        \multicolumn{3}{c}{prior works (models trained/fine-tuned on these datasets)}\\
        \midrule
        \cite{du2018harvesting} & 0.152 &  -- \\
        \cite{zhang2019addressing} & 0.184 & -- \\
        \scriptsize{UniLM Large\cite{unilmv2}} & 0.228 &  -- \\
        \scriptsize{UniLM v2 Base\cite{unilmv2}} &  0.244 & -- \\
        ERNIE-GEN Large \cite{xiao2021ernie_gen} & 0.254 & -- \\
        BART \cite{xu2022fantastic} & -- & 0.527 \\
        \midrule
        \multicolumn{3}{c}{baselines (notations defined in \S\ref{sec:problem})}\\
        \midrule
        \Mgrdy (greedy) & 0.372 & 0.424\\
        \Msmean (sample avg) & 0.359 & 0.399\\
        \Msmin (lowerbound) & 0.225 & 0.259 \\
        \Msmax (upperbound) & 0.496 & 0.548 \\
        \midrule
        \multicolumn{3}{c}{\ngram-similarity}\\
        \midrule
        uni-gram w/ context & 0.382 & 0.396 \\
        bi-gram w/ context & 0.382 & 0.403 \\
        tri-gram w/ context & 0.380 & 0.403 \\
        4-gram w/ context & 0.378 & 0.406 \\
        5-gram w/ context & 0.375 & 0.404 \\
        \midrule
        \multicolumn{3}{c}{round-trip}\\
        \midrule
        round-trip & 0.392 & 0.434 \\
        \midrule
        \multicolumn{3}{c}{prompt scores}\\
        \midrule
        grammatical correctness & 0.364 & 0.405 \\
        offensiveness & 0.374 & 0.403 \\
        clarity & 0.373 & 0.406 \\
        relevance & 0.372 & 0.396 \\
        importance & 0.372 & 0.406 \\
        specificity & 0.378 & 0.405 \\
        answerability & 0.372 & 0.404 \\
        averaged prompt score (APS) & 0.380 & 0.406 \\
        overall prompt score (OPS) & 0.373 & 0.399 \\
        \midrule
        \multicolumn{3}{c}{ensemble multiple methods}\\
        \midrule
        APS + round-trip& 0.397 & \textbf{0.439} \\
        bi-gram + round-trip & \underline{0.400} & 0.429\\
        tri-gram + round-trip & 0.398 & 0.430\\
        bi-gram + APS & 0.384 & 0.406\\
        tri-gram + APS & 0.383 & 0.409\\
        bi-gram + APS + round-trip & \textbf{0.401} & 0.431 \\
        tri-gram + APS + round-trip & \underline{0.400} & \underline{0.435} \\
        \bottomrule
    \end{tabular}
    \caption{Reference-based evaluation scores on various question selection methods. Best and second best numbers (excluding baselines) are highlighted with \textbf{boldface} and \underline{underline}, respectively.}
    \label{tab:ref_based_eval_full}
\end{table}

\begin{table*}[h!]
    \centering
    \small
    \begin{tabular}{l|cccccccc|c}
        \toprule
         & G & O & C & R & I & S & A & AHR & OHR \\
        \midrule
        GT & 0.937 & 0.987 & 0.943 & 0.930 & 0.925 & 0.922 & 0.887 & 0.933 & 0.870 \\
        \Mgrdy & 0.950 & 0.983 & 0.927 & 0.953 & 0.925 & 0.905 & 0.870 & 0.930 & 0.833 \\
        \Msmean & 0.957 & 0.981 & 0.940 & 0.937 & 0.909 & 0.921 & 0.879 & 0.932 & 0.857 \\
        \midrule
        bi-gram & 0.968 & \underline{0.990} & \textbf{0.952} & \underline{0.953} & 0.918 & \textbf{0.937} & 0.885 & 0.943 & 0.880 \\
        tri-gram & 0.968 & \underline{0.990} & 0.943 & \textbf{0.957} & \textbf{0.928} & 0.917 & 0.890 & 0.942 & 0.885 \\
        round-trip (RT) & 0.962 & \textbf{0.992} & 0.927 & 0.933 & 0.900 & 0.922 & 0.903 & 0.934 & 0.872 \\
        APS & 0.945 & 0.980 & 0.912 & 0.912 & 0.880 & 0.902 & 0.863 & 0.913 & 0.837 \\
        \midrule
        APS + RT & 0.947 & 0.977 & 0.928 & 0.915 & 0.890 & 0.898 & 0.903 & 0.923 & 0.875 \\
        bi-gram + RT & \textbf{0.983} & 0.980 & \textbf{0.952} & 0.950 & 0.920 & \textbf{0.937} & \underline{0.917} & \underline{0.948} & \underline{0.893} \\
        tri-gram + RT & \textbf{0.983} & 0.987 & \textbf{0.952} & \underline{0.953} & \underline{0.927} & \underline{0.933} & \textbf{0.925} & \textbf{0.951} & \textbf{0.900} \\
        bi-gram + APS & 0.972 & 0.983 & 0.943 & 0.932 & 0.903 & 0.922 & 0.867 & 0.932 & 0.857 \\
        tri-gram + APS & 0.972 & 0.983 & 0.945 & 0.928 & 0.910 & 0.912 & 0.878 & 0.933 & 0.862 \\
        bi-gram + APS + RT & \underline{0.977} & 0.980 & 0.947 & 0.932 & 0.905 & 0.918 & 0.905 & 0.938 & 0.877 \\
        tri-gram + APS + RT & 0.972 & 0.980 & \underline{0.948} & 0.928 & 0.910 & 0.912 & 0.902 & 0.936 & 0.878 \\
        \bottomrule
    \end{tabular}
    \caption{Human eval results (SQuAD). Abbreviations in the first row denote \textbf{G}rammatical correctness, \textbf{O}ffensiveness, \textbf{C}larity, \textbf{R}elevance, \textbf{I}mportance, \textbf{S}pecificity, \textbf{A}nswerability, \textbf{A}veraged \textbf{H}uman \textbf{R}ating (over all dimensions to the left), \textbf{O}verall \textbf{H}uman \textbf{R}ating (an overall score given by annotators). Best and second best numbers (excluding baselines) are highlighted with \textbf{boldface} and \underline{underline}, respectively.}
    \label{tab:human_squad}
\end{table*}

\begin{table*}[h!]
    \centering
    \small
    \begin{tabular}{l|cccccccc|c}
        \toprule
         & G & O & C & R & I & S & A & AHR & OHR \\
        \midrule
        GT & 0.945 & 0.963 & 0.942 & 0.937 & 0.885 & 0.928 & 0.892 & 0.927 & 0.867 \\
        \Mgrdy & 0.975 & 1.000 & 0.958 & 0.943 & 0.920 & 0.922 & 0.905 & 0.946 & 0.870 \\
        \Msmean & 0.964 & 0.988 & 0.944 & 0.955 & 0.925 & 0.934 & 0.912 & 0.946 & 0.875 \\
        \midrule
        bi-gram & \textbf{0.953} & \textbf{0.993} & 0.943 & 0.943 & \textbf{0.932} & 0.937 & 0.902 & 0.943 & 0.857 \\
        tri-gram & 0.943 & 0.980 & 0.922 & 0.930 & 0.905 & 0.927 & 0.858 & 0.924 & 0.838 \\
        round-trip (RT) & 0.928 & 0.970 & 0.945 & 0.937 & 0.888 & 0.935 & 0.878 & 0.926 & 0.862 \\
        APS & \underline{0.952} & 0.985 & 0.957 & \underline{0.972} & 0.922 & \textbf{0.977} & \textbf{0.948} & \textbf{0.959} & \textbf{0.895} \\
        \midrule
        APS + RT & 0.927 & 0.983 & 0.955 & \textbf{0.973} & 0.915 & \underline{0.948} & 0.928 & 0.947 & \underline{0.893} \\
        bi-gram + RT & 0.943 & \underline{0.992} & 0.952 & 0.938 & 0.910 & 0.927 & 0.918 & 0.940 & 0.860 \\
        tri-gram + RT & 0.927 & 0.973 & 0.932 & 0.928 & 0.890 & 0.935 & 0.882 & 0.924 & 0.857 \\
        bi-gram + APS & 0.945 & 0.988 & \textbf{0.962} & 0.942 & \underline{0.925} & 0.945 & \underline{0.932} & \underline{0.948} & 0.888 \\
        tri-gram + APS & 0.948 & 0.978 & 0.952 & 0.938 & 0.918 & 0.940 & 0.913 & 0.941 & 0.863 \\
        bi-gram + APS + RT & 0.945 & 0.988 & \underline{0.960} & 0.942 & 0.913 & 0.937 & \underline{0.932} & 0.945 & 0.883\\
        tri-gram + APS + RT & 0.925 & 0.987 & 0.938 & 0.962 & 0.910 & 0.942 & 0.922 & 0.941 & 0.882 \\
        \bottomrule
    \end{tabular}
    \caption{Human eval results (Fairytale QA). Abbreviations in the first row denote \textbf{G}rammatical correctness, \textbf{O}ffensiveness, \textbf{C}larity, \textbf{R}elevance, \textbf{I}mportance, \textbf{S}pecificity, \textbf{A}nswerability, \textbf{A}veraged \textbf{H}uman \textbf{R}ating (over all dimensions to the left), \textbf{O}verall \textbf{H}uman \textbf{R}ating (an overall score given by annotators). Best and second best numbers (excluding baselines) are highlighted with \textbf{boldface} and \underline{underline}, respectively.}
    \label{tab:human_ft}
\end{table*}

\section{Implementation Details}
\label{sec:app:implementation}

In all experiments, we use the \token{text-davinci-002} (175B parameters) variant of GPT-3. 
It is currently the most capable GPT-3 model variant.
Compared to other variants, \token{text-davinci-002}'s support to inserting completions can better facilitate our question generation tasks (as shown in Figure~\ref{fig:prompt_qgen}).

We use a temperature of 0.7 during the sampling process of question generation.
In all other use cases (e.g., QA round-trip, prompt score), we use greedy generation (temperature is set to 0).

\end{document}